\documentclass{article}

\usepackage{graphicx} 
\usepackage{multirow}

\usepackage[final]{neurips_2025}

\usepackage[utf8]{inputenc} 
\usepackage[T1]{fontenc}    
\usepackage{hyperref}       
\usepackage{url}            
\usepackage{booktabs}       
\usepackage{amsfonts}       
\usepackage{nicefrac}       
\usepackage{microtype}      
\usepackage{xcolor}         
\usepackage{hyperref}
\usepackage{ragged2e}       
\usepackage{tabularx,ragged2e} 

\title{Position: Thematic Analysis of Unstructured Clinical Transcripts with Large Language Models}

\author{
Seungjun Yi\textsuperscript{1\thanks{Corresponding author}},
Joakim Nguyen\textsuperscript{1},
Terence Lim\textsuperscript{1},
Andrew Well\textsuperscript{2},
Joseph Skrovan\textsuperscript{1},\\
\textbf{Mehak Beri\textsuperscript{1},
YongGeon Lee\textsuperscript{1},
Kavita Radhakrishnan\textsuperscript{1},
Liu Leqi\textsuperscript{1},
Mia Markey\textsuperscript{1},
Ying Ding\textsuperscript{1}} \\
\textsuperscript{1}The University of Texas at Austin, Austin, TX, USA\\
\textsuperscript{2}Department of Cardiac Surgery, Vanderbilt University School of Medicine, Nashville, TN, USA\\
\texttt{\{charlie.yi,jhn001,terence.lim\}@utexas.edu},\\
\texttt{andrew.well@vumc.org},\,
\texttt{jskrovan@fastmail.fm},\
\texttt{mb72384@eid.utexas.edu},\\,\texttt{\{yg910524\}@utexas.edu},\,
\texttt{kradhakrishnan@mail.nur.utexas.edu},\\
\texttt{\{leqiliu,mia.markey\}@utexas.edu},\,
\texttt{ying.ding@ischool.utexas.edu}
}

\begin{document}

\maketitle

\begin{abstract}
This position paper examines how large language models (LLMs) can support thematic analysis of unstructured clinical transcripts, a widely used but resource-intensive method for uncovering patterns in patient and provider narratives. We conducted a systematic review of recent studies applying LLMs to thematic analysis, complemented by an interview with a practicing clinician. Our findings reveal that current approaches remain fragmented across multiple dimensions including types of thematic analysis, datasets, prompting strategies and models used, most notably in evaluation. Existing evaluation methods vary widely (from qualitative expert review to automatic similarity metrics), hindering progress and preventing meaningful benchmarking across studies. We argue that establishing standardized evaluation practices is critical for advancing the field. To this end, we propose an evaluation framework centered on three dimensions: \emph{validity}, \emph{reliability}, and \emph{interpretability}. 
\end{abstract}







\section{Introduction}
Thematic analysis (TA) is one of the most widely used methods for qualitative data analysis, providing a structured yet flexible framework for identifying, analyzing, and reporting patterns (\textit{themes\footnote{\textit{Themes}: Broader patterns or categories that emerge from grouping related codes to capture significant aspects of participants’ experiences or perspectives in relation to the research objectives}}) within textual data~\cite{braun2006thematic,clarke2013teaching}. In clinical contexts, TA is frequently applied to unstructured patient interview transcripts to reveal underlying meanings and generate actionable insights for improving care~\cite{AHMED2025100198}, with the process commonly guided by Braun and Clarke’s six-phase framework~\cite{braun2006thematic}.
Researchers begin by familiarizing themselves with the data through repeated readings and the identification of important keywords (Step 1–2). These keywords are then used to generate initial \textit{codes}\footnote{\textit{Codes:} concise labels that identify features of the textual data relevant to the research question} (Step 3), which serve as the foundation for identifying preliminary \textit{themes} (Step 4). The \textit{themes} are iteratively reviewed, refined, and clearly defined and named through interpretive analysis (Step 5). In the final stage, a structured conceptual model is developed to represent the thematic relationships and underlying constructs (Step 6). To ensure credibility, at least two trained researchers with domain expertise must independently engage with the data and participate in multiple rounds of discussion to achieve consensus.
Each year, over 900,000 U.S. healthcare interviews undergo thematic analysis across academic, clinical, and research settings~\cite{dejonckheere2019semistructured, vasileiou2018sample}. Most involve unstructured data, requiring inductive thematic analysis (ITA), which extracts \textit{themes} directly from transcripts without predefined \textit{codes}\cite{campbell2021reflexive, taylor2018rapid, watkins2017qualitative}. Manual ITA demands over 6.1 million hours annually equivalent to 3,000 full-time jobs and \$305.4 million in cost which limits scalability and delaying insights in fast-moving healthcare environments~\cite{namey2008data, Hamza2014coding, hennink2017saturation}. Automating ITA is therefore critical for timely, scalable, and reproducible \textit{theme} generation.


Recent advances in large language models (LLMs) offer a promising alternative for automating key components of the six-step manual inductive thematic analysis process~\citep{yi2025protomedllmautomaticevaluationframework, yi2025autota}. LLMs can rapidly generate initial \textit{codes} (Step 2) and candidate \textit{themes} (Steps 3–5) from interview transcripts in under 10 minutes, a substantial reduction compared to the 5–8 hours required by human analysts~\cite{xu2025tamahumanaicollaborativethematic, raza2025llmtallmenhancedthematicanalysis}.

However, the integration of LLMs to fully automate the TA process remains in its early stages.
Existing studies primarily focus on specific stages of thematic analysis, such as \textit{coding}~\citep{parfenova-etal-2025-text}, and \textit{theme} generation~\citep{Mannstadt2024Novel, deiner2024llmita}.
They also vary widely in methodological choices, including the type of thematic analysis employed (i.e., inductive vs.~deductive), the models used (e.g., \texttt{gpt-4o}, \texttt{Llama}), and the prompting strategies adopted (e.g., few-shot prompting, chain-of-thought), which makes it difficult to establish a coherent picture of progress. This challenge is further compounded by the rapid pace of LLM development, which continuously reshapes the methodological landscape. 
More importantly, evaluating the results of thematic analysis remains a major challenge. Evaluation practices remain inconsistent, relying primarily on expert qualitative assessment and comparisons between LLM-generated and human-generated themes, which limits both reproducibility and comparability across studies.

In this work, we aim to:
\begin{enumerate}
    \item Provide an overview of recent developments and research progress in thematic analysis supported by large language models (LLMs).
    \item \textbf{Highlight the need for standardized evaluation in LLM-based thematic analysis, and propose a evaluation framework based on validity, reliability, and interpretability.}
    \item Emphasize the need for end-to-end frameworks to ensure the practical utility of thematic analysis in clinical settings.
\end{enumerate}

\section{Methods}

\paragraph{Literature Search and Selection}
\label{sec:data-collection}

Our data collection targeted publications addressing thematic analysis (TA) in relation to large language models (LLMs). Using arXiv metadata (cutoff: 2025–08–15), we identified relevant papers through keyword-based searches of titles and abstracts. Summary statistics are reported in Table\ref{tab:ta_llm_stats}, with full search details in the Appendix~\ref{app:search-details}.
To address the limitation that some peer-reviewed papers may be missing from arXiv, we supplemented our search with Elicit~\citep{kung2023elicit} and manual queries on major indexing platforms (see Appendix~\ref{app:supplementary-search} for details). Full procedures, verification steps, survey prompts, and keyword lists are provided in Appendix~\ref{app:survey_prompts} and Table~\ref{tab:keywords_dict}.
From the collected literature, we reviewed 56 recent (past three years from the cutoff date) studies applying LLMs to thematic analysis, focusing on five dimensions: analysis type, models, datasets, prompting strategies, and evaluation methods. 
We included peer-reviewed publications, preprints, while excluding theses, dissertations, and incomplete work.

\paragraph{Expert Interview}
In addition to the systematic literature review, we conducted an in-depth interview with a practicing cardiac surgeon with eight years of clinical experience, who has previously performed manual TA of patient transcripts~\cite{mery2023journey}. The interview was held over a two-hour Zoom session, during which the surgeon was provided with a summary of the literature review findings and invited to discuss both their implications and the broader role of  LLMs in augmenting thematic analysis workflows. Detailed insights from this discussion are presented in Section~\ref{sec:discussion-conclusion}.

\section{Results}

\subsection{Trends and Dataset Characteristics of LLM-Supported Thematic Analysis (TA)}

\begin{figure}[h]
    \centering
    \includegraphics[width=\linewidth]{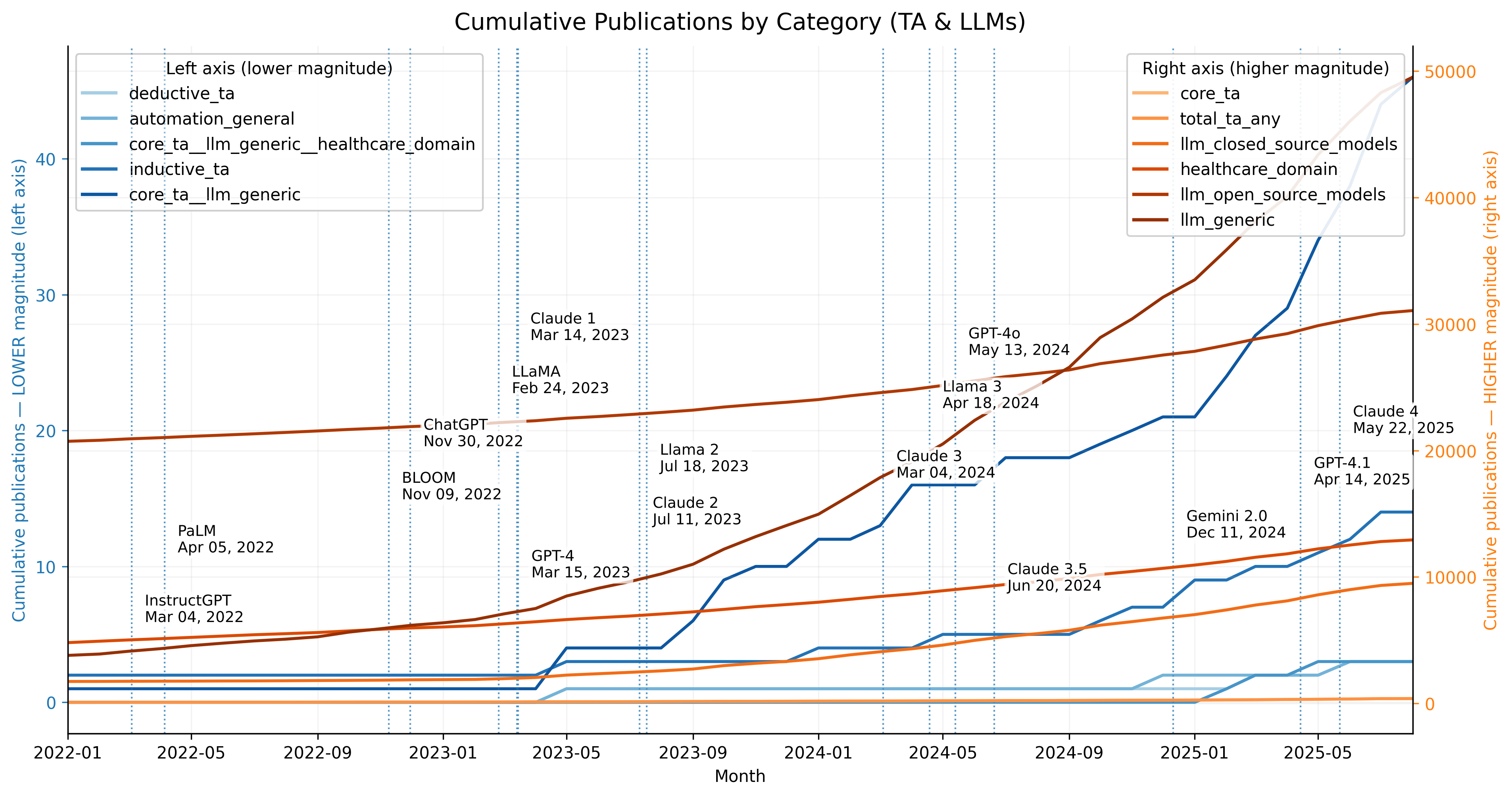}
    \caption{Cumulative number of publications mentioning thematic analysis (TA) and LLMs by category since Jan 2022. Research explicitly combining TA with LLMs began to emerge following the release of GPT-4. Vertical lines indicate major milestones in LLM development (see Appendix~\ref{app:milestones}). The categories shown correspond to those summarized in Table~\ref{tab:ta_llm_stats}.}
    \label{fig:one}
\end{figure}

From Table~\ref{tab:ta_llm_stats}, we observe that TA leveraging LLMs is a relatively recent phenomenon, with the first work appearing after the release of GPT-3.5 in Nov 2022~\cite{dai2023llmintheloopleveraginglargelanguage}. The number has been growing exponentially since then (Figure~\ref{fig:one}). As of Aug 2025, total of 46 papers were identified in the arXiv metadata, which only three were related to healthcare. 

\begin{table}[h]
\centering
\caption{Summary of retrieved papers containing keywords related to thematic analysis and LLMs. 
The additional ``(+11)'' indicates papers identified outside of arXiv. 
Keyword list definitions are provided in Appendix Table~\ref{tab:keywords_dict}.}
\label{tab:ta_llm_stats}
\begin{tabular}{l r}
\toprule
\textbf{Keyword Category} & \textbf{Count} \\
\midrule
Thematic Analysis (\texttt{core\_ta}) & 381 \\
Inductive Thematic Analysis (\texttt{inductive\_ta}) & 14 \\
Deductive Thematic Analysis (\texttt{deductive\_ta}) & 3 \\
Thematic Analysis (\texttt{core\_ta}) + LLMs (\texttt{llm\_generic}) & 45 (+11)\\
Thematic Analysis (\texttt{core\_ta}) + LLMs (\texttt{llm\_generic}) + Healthcare (\texttt{healthcare}) & 3 \\
Total TA (any category above) & 428 \\
\bottomrule
\end{tabular}
\end{table}

\newcolumntype{Y}{>{\RaggedRight\arraybackslash}X} 



\begin{table}[ht]
\centering
\small
\caption{Frequency summary across dimensions for 56~papers (Aug 2022–Aug 2025). Since individual papers may span multiple categories, counts can exceed 56. Categories deemed not applicable were excluded from the table.}
\label{tab:freq_multi_dimension}
\begin{tabularx}{\linewidth}{p{2.8cm} Y}
\toprule
\textbf{Dimension} & \textbf{Attributes (\# out of 56 / \%)} \\
\midrule

Types of TA &
Inductive (36/64\%), Few-shot (9/16\%), Hybrid (12/22\%), Other (3/5\%) \\

Models (by family) &
GPT (32/58\%), Claude (7/13\%), LLaMA (6/11\%), Gemini (6/11\%), 
\mbox{Mistral/Mixtral (3/5\%)}, DeepSeek (4/7\%), Qwen (2/4\%), 
Other/Custom (7/13\%) \\

Datasets by Domain &
Social Media/Online Communities (14/25\%), Education (12/21\%), 
\mbox{Software Engineering/Programming (11/20\%)}, Arts/Humanities/Other (10/18\%), 
Healthcare/Clinical (9/16\%) \\

Prompting &
Zero-shot (19/35\%), Few-shot (9/16\%), Chain-of-thought (7/13\%), 
\mbox{Self-consistency/Reflexion (8/15\%)}, Tool/Agent use (7/11\%) \\

Evaluation &
Human qualitative review (22/40\%), Automatic text metrics (16/27\%), 
\mbox{Task-based/utility eval (7/13\%)}, Hybrid (11/20\%) \\
\bottomrule
\end{tabularx}
\end{table}

\subsection{Survey of LLM-Assisted Thematic Analysis (TA)}
We explored five key dimensions of LLM-assisted TA, including types of TA adopted, models leveraged, datasets analyzed, prompting strategies employed, and evaluation methods applied (Table~\ref{tab:freq_multi_dimension}). 


\paragraph{Type of Thematic Analysis: Inductive vs.~Deductive.}

 \emph{Deductive} TA, grounded in a priori codebooks, provides clearer definitions and comparability, whereas \emph{inductive} TA generates themes directly from transcripts, enabling the derivation of novel insights.
Among the 56 studies, inductive TA was the most prevalent, accounting for 36 papers (64\%). Hybrid approaches that combined inductive and deductive elements were the second most common (22\%), while purely deductive analyses were relatively rare (9\%). A small fraction of work (5\%) employed other or less clearly specified strategies. 
\emph{Deductive} and \emph{inductive} TA require different evaluation approaches. Consequently, evaluation methods developed for one are difficult to transfer to the other, which contributes to persistent inconsistency in evaluation practices.

\paragraph{Models Employed.}  

Across the surveyed literature, the majority of studies relied on the OpenAI GPT family, with GPT-3.5~\cite{brown2020languagemodelsfewshotlearners} and GPT-4~\cite{openai2024gpt4technicalreport} variants serving as the primary analytic engines. A smaller number incorporated Claude~\cite{anthropic2023claude} or Gemini~\cite{gemini2023} models, most often as comparative baselines rather than central analytic tools. Open-source families such as LLaMA~\cite{touvron2023llamaopenefficientfoundation}, Mistral~\cite{jiang2023mistral7b}, and DeepSeek~\cite{deepseek2025} appeared in exploratory or ensemble configurations, reflecting growing interest in transparency and cost-efficient alternatives. Only a handful of papers examined fine-tuned or specialized deployments (e.g., InCoder~\cite{fried2023incodergenerativemodelcode}, CodeGen~\cite{nijkamp2023codegenopenlargelanguage}, ProgGP~\cite{austin2021programsynthesislargelanguage}), typically for narrow use cases. A detailed breakdown by model family is provided in Appendix~\ref{app:models-employed}.

\paragraph{Datasets by Domain.}  
The surveyed studies drew on diverse domains, with social media and online communities~\cite{ghali2025beyondwordsneedagenticgenerative, qiao2025thematiclm, hairston2025automatedthematicanalysesusing} most common (25\%), followed by education~\cite{kazemitabaar2023novicesusellmbasedcode, harvey2025dontforgetteacherseducatorcentered} (21\%), software engineering and programming~\cite{kazemitabaar2023novicesusellmbasedcode, Kazemitabaar_2024} (20\%), and healthcare or clinical contexts~\cite{yi2025autota, raza2025llmtallmenhancedthematicanalysis, Mathis2024ITA}
 (16\%). The remaining 18\% were arts, humanities, and other fields such as court cases, disciplinary decisions, and design workshops. 



\paragraph{Prompting Strategies.}  

Studies most commonly relied on zero-shot prompting (i.e., with no specific examples given), often implemented through multi-stage pipelines, with smaller proportions using few-shot templates, chain-of-thought reasoning, or iterative refinement methods such as self-consistency and Reflexion. A minority explored multi-agent or tool-based prompting, while some did not employ prompting at all, focusing instead on fine-tuned or human-coded approaches. Overall, zero-shot prompting dominated for its accessibility and alignment with inductive thematic analysis, whereas more elaborate prompting designs reflected attempts to stabilize outputs or approximate collaborative coding practices. A fuller breakdown is provided in Appendix~\ref{app:prompting}.
Also, most existing approaches still rely on human-in-the-loop workflows that require full transcript review, which severely limits scalability. Since familiarization (Step 1 of Braun \& Clarke’s six-step thematic analysis process \cite{braun2006thematic}) with transcripts is the most time-consuming step, retaining this requirement is essential to the practical utility of LLM-assisted thematic analysis in clinical settings. 

\paragraph{Evaluation.}  

Evaluation of LLM-assisted thematic analysis remains fragmented, with no universally accepted gold standard. 
Human qualitative review was the most common approach ($\approx40\%$), involving inter-coder agreement checks, reflexive discussions, and expert evaluations of plausibility and thematic coverage, although reproducibility was often limited. Inter-coder agreement was typically assessed using Krippendorff’s alpha~\cite{krippendorff2004reliability}, a reliability coefficient that accounts for chance agreement, supports different levels of measurement, and handles incomplete data, or Cohen’s kappa~\cite{cohen1960coefficient}, which measures agreement between two raters under the assumption of complete data and equal category distribution. 
Other criteria beyond inter-coder agreement were measured through surveys in which experts rated on a 1–5 Likert scale~\citep{raza2025llmtallmenhancedthematicanalysis, parfenova-etal-2025-text}. 
About one quarter of studies (27\%) evaluated LLM-generated themes against human ground truth using automatic similarity metrics. Lexical overlap metrics such as Jaccard Index, fuzzy match,  ROUGE~\cite{lin2004rouge}, BLEU~\cite{papineni2002bleu}, GLUE~\cite{wang2019gluemultitaskbenchmarkanalysis}, and METEOR~\cite{lavie-agarwal-2007-meteor} assess similarity based on shared words or n-grams\footnote{Overlapping sequences of one or more consecutive words.}. Semantic similarity metrics, including  cosine similarity, BERTScore~\cite{zhang2020bertscoreevaluatingtextgeneration}, compare vector embeddings\footnote{Numerical representations of text that capture semantic meaning, generated by models such as \texttt{all-mpnet-base-v2} and \texttt{text-embedding-3-large}.}. Classifier-based accuracy and F1 scores were occasionally applied.
A smaller subset (13\%) evaluated outputs indirectly through task-based measures (e.g., learning outcomes), 
while about one fifth (20\%) adopted hybrid designs combining human judgments with computational metrics described above~\cite{parfenova-etal-2025-text}.

However, major challenges in evaluation remain.  
Even when the same metric is reported, differences in underlying embeddings hinder comparability and synthesis across studies. Issues of transparency further compound these challenges: the surveyed work do not disclose the complete ground truth, often citing privacy or IRB restrictions, while others provide no explicit justification for withholding this information.
For the usage of automatic metrics, a major limitation of directly assessing similarity between human- and LLM-generated themes is the lack of one-to-one mappings. Thematic analysis inherently produces sets of themes that only partially overlap: a single LLM-generated theme may correspond to multiple human themes, while some human themes may have no direct counterpart.  

To address these limitations, we propose evaluating automated inductive thematic analysis along three key dimensions: \textbf{validity}, \textbf{reliability}, and \textbf{interpretability}.
(1) \textbf{Validity.}
Any set of generated themes may not respect a 1:1 mapping between humans and LLMs, so we compute \textbf{maximum-weight bipartite matching} on a similarity matrix $\mathbf{S}$, where $S_{ij}$ denotes the similarity between human theme $i$ and model theme $j$.
Use both:
(a) Lexical overlap: Jaccard; optionally ROUGE for n-gram overlap. (b) Semantic similarity: cosine over sentence embeddings or BERTScore / MoverScore / BLEURT~\citep{zhang2020bertscoreevaluatingtextgeneration}, and 
    report: Precision/Recall/F1@match, Coverage@$\tau$ (fraction of human themes matched above a threshold), \emph{Redundancy} (mean intra-LLM theme similarity), and \emph{Novelty rate} (LLM themes with no matched human counterpart). 
(2) \textbf{Reliability.}
Krippendorff’s $\alpha$ or Cohen/Fleiss $\kappa$ can be used as diagnostics for vague definitions or unclear code boundaries. To test \emph{stability}, re-run parts of the pipeline with different random seeds or bootstrap samples and compute Adjusted Rand Index (ARI) or Variation of Information (VI) for code assignments. At the theme level, system-generated themes can be aligned with human themes by comparing overlap in supporting excerpts. When applying LLMs to the pipeline, \emph{confirmability} assesses whether themes are data-driven or driven by the internal biases inherent across different LLMs.
(3) \textbf{Interpretability.}
Analytical depth and nuance remain challenging~\citep{parkington2025humanvsllmbasedthematic}. Mitigations include rationale generation~\citep{dunivin2024scalablequalitativecodingllms}, expert adjudication~\citep{raza2025llmtallmenhancedthematicanalysis}, and adaptive codebooks~\citep{qiao2025thematiclm}. Excerpts within a theme should be consistent in meaning, and different
themes should be distinguishable. Embedding similarity can check \emph{coherence} defined as the average similarity among quotes within each theme and \emph{distinctiveness} as the distance between theme centroids~\citep{calinski1974vrc}. Report \emph{coverage} of (i) passages assigned to any theme and (ii) participants represented by each theme. The \emph{credibility} of themes assesses whether the generated themes faithfully represent the data. For domain-specific contexts, combine automated metrics with human-in-the-loop validation~\citep{bedemariam2025potentialperilslargelanguage,hairston2025automatedthematicanalysesusing}.

\paragraph{Other considerations: Cost of Thematic Analysis.} 
The cost of using LLMs has become significantly more feasible. Recent analyses estimate inference prices around \$0.15–\$2.50 USD per million input tokens and \$0.60–\$10 USD per million output tokens, compared to the \$200{,}000+ typically required for manual thematic analysis~\citep{erol2025costofpasseconomicframeworkevaluating, Bergemann2025EconomicsLLM}.
\label{app:cost}

\section{Discussion and Conclusion}
\label{sec:discussion-conclusion}



Our results indicate that current LLM-based approaches to thematic analysis remain fragmented, especially in the evaluation methods used across studies. As the clinician in our study interview observed, evaluating inductive thematic analysis can feel like ``\textit{shooting in the dark},'' given the absence of clear ground truths or universally accepted metrics. 
The current disparate evaluation practices prevents direct comparison across studies. Without such standardization, research efforts remain fragmented and the development of an end-to-end automated thematic analysis framework with LLMs is hindered.

In summary, while LLMs hold considerable potential to transform thematic analysis, progress will remain fragmented without common standards for evaluation. By advancing validity, reliability, and interpretability as shared dimensions, the field can move toward more rigorous, scalable, and clinically meaningful applications of automated qualitative research.

\clearpage
\bibliographystyle{unsrt}
\bibliography{references}


\appendix
\section{Keyword Dictionary}

\begin{table}[h]
\centering
\caption{Keyword dictionary (\texttt{KEYWORDS\_DICT}) used for thematic analysis literature search.}
\label{tab:keywords_dict}
\begin{tabular}{p{4cm} p{9.5cm}}
\toprule
\textbf{Dictionary Key} & \textbf{Keywords} \\
\midrule
\texttt{core\_ta} & thematic analysis; qualitative thematic analysis; theme analysis; thematic coding; theme coding; identify themes; theme identification; Braun and Clarke; Braun \& Clarke; six-phase framework; six phase framework \\
\texttt{inductive\_ta} & inductive thematic analysis; data-driven thematic analysis; bottom-up thematic analysis; open coding (qualitative) \\
\texttt{deductive\_ta} & deductive thematic analysis; theory-driven thematic analysis; top-down thematic analysis; a priori codes \\
\texttt{llm\_generic} & LLM; large language model; transformer-based model; foundation model; prompting; chain-of-thought; self-reflection (LLM); RAG; few-shot; zero-shot \\
\texttt{healthcare} & clinical interview; patient interview; healthcare; medicine; clinical transcript; medical transcript; qualitative health research; nursing \\
\bottomrule
\end{tabular}
\end{table}

\section{Model Release Milestones}
\label{app:milestones}
We identify the following milestones in the development of LLMs; however, depending on which models were highlighted in the literature review, only a subset of these key milestones were selected for inclusion in our plot. InstructGPT was introduced on March 4, 2022, followed by PaLM on April 5, 2022, and BLOOM on November 9, 2022. ChatGPT launched on November 30, 2022. In 2023, LLaMA appeared on February 24, Claude 1 on March 14, GPT-4 on March 15, Claude 2 on July 11, and Llama 2 on July 18. The year 2024 brought Claude 3 on March 4, Llama 3 on April 18, GPT-4o on May 13, Claude 3.5 on June 20, and Gemini 2.0 on December 11. In 2025, GPT-4.1 was released on April 14, followed by Claude 4 on May 22, Claude 4.1 on August 5, and GPT-5 on August 7.

\section{Data Collection and Keyword Search Procedure}
\label{app:search-details}

We accessed the official arXiv metadata via the \href{https://www.kaggle.com/datasets/Cornell-University/arxiv}{Kaggle API} (cutoff date: 2025–08–15). Each record included fields such as title, abstract, and subject categories.
\begin{itemize}
    \item \textbf{Keyword construction:} We compiled keyword lists for thematic analysis (core TA, inductive TA, deductive TA), large language models (LLMs), and the healthcare domain (Appendix Table~\ref{tab:keywords_dict}).
    \item \textbf{Search method:} A case-insensitive string search was applied to the title and abstract fields. A paper was counted in a category if it contained at least one keyword from that list. Multiple matches within a category were counted once. For example, papers with several TA terms were recorded once under \texttt{core\_ta}.
    \item \textbf{Category intersections:} Papers with both TA and LLM terms were classified under \texttt{core\_ta} + \texttt{llm\_generic}. Those additionally containing healthcare terms were counted in the three-way intersection (\texttt{core\_ta} + \texttt{llm\_generic} + \texttt{healthcare}).
\end{itemize}

The resulting counts are summarized in Table~\ref{tab:ta_llm_stats}.

\section{Survey Prompts}
\label{app:survey_prompts}

The prompt used for the literature search in \textit{Elicit}~\cite{kung2023elicit} was as following:
\textit{How can evaluation frameworks for automated inductive thematic analysis balance validity, reliability, and interpretability when comparing direct LLM coding and hybrid human–AI workflows?}

\section{Supplementary Literature Search Strategies}
\label{app:supplementary-search}

To complement the arXiv-based search and acknowledge the limitation that some peer-reviewed papers may be missing from arXiv (as it is a non-archival venue), we implemented two additional strategies:

\begin{itemize}
    \item \textbf{Elicit-assisted search:} We used Elicit~\citep{kung2023elicit}, an AI-powered research assistant designed to streamline the literature review process. All references identified through Elicit were manually verified and reviewed by the authors. Specific examples of the survey prompts are provided in Appendix~\ref{app:survey_prompts}.
    \item \textbf{Manual database search:} We conducted manual searches on major indexing platforms\footnote{\href{https://scholar.google.com/}{Google Scholar}, \href{https://pubmed.ncbi.nlm.nih.gov}{PubMed}, \href{https://www.scopus.com}{Scopus}, \href{https://www.webofscience.com}{Web of Science}} using the query \textit{``thematic analysis''} in combination with domain-specific terms. The exact keyword lists are reported in Appendix Table~\ref{tab:keywords_dict}.
\end{itemize}

\section{Models Employed}
\label{app:models-employed}

\noindent \textbf{OpenAI GPT family (41 papers, $\sim$74\%)}:  
GPT-3.5 and GPT-4 variants (Turbo, 16k/32k, GPT-4o, GPT-4V). Backbone of most studies.  
\noindent \textbf{Claude (8 papers, $\sim$15\%)}:  
Claude-instant, Claude-2/3.5/3.7 Sonnet. Used mainly for comparison.
\noindent \textbf{LLaMA \& open-source (7 papers, $\sim$13\%)}:  
LLaMA-2/3, CodeLlama, Mistral, Mixtral, Gemma, DeepSeek. Benchmarked against GPT; valued for transparency.  
\noindent \textbf{Google Gemini (6 papers, $\sim$11\%)}:  
Gemini Pro/Flash/Ultra/2.5 Preview. Typically baseline; mixed performance on inductive tasks.  
\noindent \textbf{Other (5 papers, $\sim$9\%)}:  
InCoder, CodeGen, ProgGP, genre-CTRL, Whisper, Sentence-T5. Niche or auxiliary roles.  


\section{Prompting Strategies}
\label{app:prompting}

\noindent \textbf{Zero-shot (35\%)}: Most common; single- or multi-stage pipelines for code extraction, theme consolidation, and naming. Efficient but sensitive to phrasing, raising reproducibility concerns.  
\noindent \textbf{Few-shot (16\%)}: Exemplars, templates, or pseudo-code scaffolds embedded in prompts; improved controllability but introduced bias in exemplar selection.  
\noindent \textbf{Chain-of-thought (13\%)}: Encouraged intermediate reasoning steps; supported transparency, though outputs were often verbose or inconsistent.  
\noindent \textbf{Self-consistency/Reflexion (15\%)}: Iterative or recursive re-prompting; aimed at stability but benefits over single-pass outputs were mixed.  
\noindent \textbf{Tool/agent-based (11\%)}: Multi-agent roles (coder, evaluator, judge) or debate frameworks; innovative but resource-intensive, with variable gains.  
\noindent \textbf{Not applicable (11\%)}: No prompting used (e.g., fine-tuned models, human-coded datasets, conceptual papers).  



\end{document}